\documentclass[runningheads]{llncs}

\usepackage{eccv}

\usepackage{eccvabbrv}

\usepackage{comment}

\usepackage{graphicx}
\usepackage{multirow}
\usepackage{booktabs}
\usepackage{subcaption}

\usepackage[accsupp]{axessibility}  

\usepackage{hyperref}

\usepackage{orcidlink}

\begin{document}

\title{Robust Validation to Geometric Perturbations for Autonomous Pose Estimation}

\titlerunning{Robust Validation to Geometric Perturbations}

\author{Gregoire Theau\inst{1} \and
Melanie Ducoffe\inst{1,2}}

\authorrunning{G.~Theau and M.~Ducoffe}

\institute{Airbus SAS, France \and
IRT Saint-Exupery, France\\
}

\maketitle

\begin{abstract}
Deploying autonomous systems in safety-critical domains demands guaranteed robustness against physically plausible geometric perturbations rather than abstract pixel-wise noise. In vision-based navigation and autonomous landing, machine learning components require rigorous validation under dynamic operational conditions such as camera rotations and lighting shifts. Extending findings on the failure of first-order spatial attacks in classification, we show that standard gradient-based heuristics (e.g. APGD) similarly fail on for pose estimation, often performing worse than a simple random sampling baseline. 

To overcome these optimization bottlenecks, we reformulate pose estimation robustness within the framework of Global Lipschitzian Optimization (GLO). We argue that GLO offers a principled approach to robust validation, effectively localizing global optima with strong theoretical convergence guarantees. We evaluate this framework on a YOLOv8-Pose keypoint detector with a Perspective-n-Point (PnP) solver against rotation and contrast. In our evaluations, GLO successfully isolates critical failure modes where position deviations exceed safe operational limits, while rapidly pruning the search space by over $80\%$. To the best of our knowledge, this is the first study to extend geometric robustness validation to continuous keypoint regression and deep object detection, establishing a practical step toward certifying robust autonomous perception.

\end{abstract}

\begin{figure}[t]
    \centering
    \includegraphics[width=0.8\textwidth]{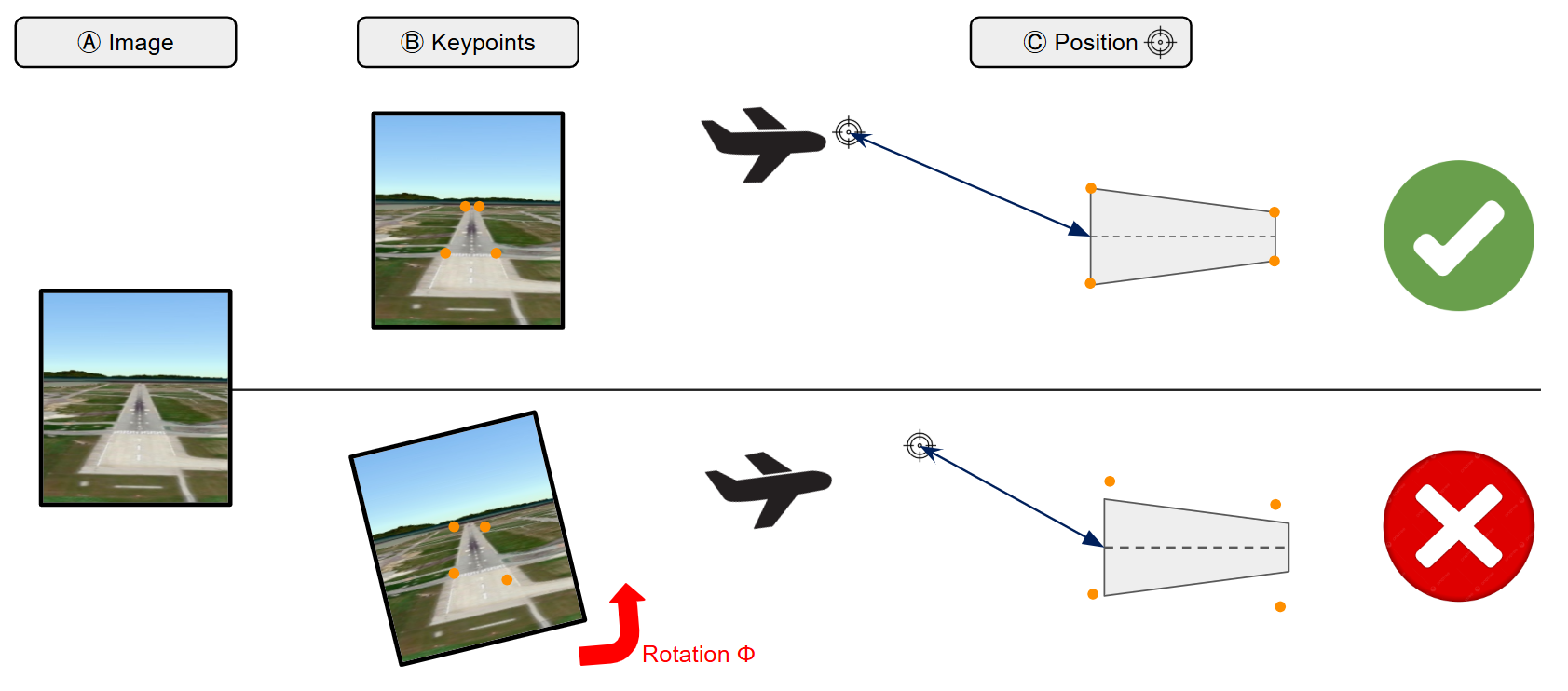}
    \caption{Overview of the Vision-Based Landing (VBL) robustness validation framework. \textbf{(A) Image:} The input image, subjected to physically plausible geometric perturbations such as camera rotation. \textbf{(B) Keypoints:} The 2D runway corners extracted by the YOLOv8-Pose detector. \textbf{(C) Position:} The 3D aircraft pose estimated by the Perspective-n-Point (PnP) solver. The end-to-end pipeline is evaluated using Global Lipschitzian Optimization (GLO) to isolate critical failure modes.}
    \label{fig:overview_pipeline}
\end{figure}
\section{Introduction}

Deploying deep neural networks in safety-critical autonomous systems demands guaranteed robustness against physically plausible operational disturbances rather than abstract pixel-wise noise \cite{szegedy2014intriguing}. Perception chains in self-driving vehicles, mobile robotics, and aerospace serve as the critical bridge between raw visual inputs and trajectory control. While deep vision models achieve high benchmark accuracy, their vulnerability to physical distribution shifts poses severe risks to closed-loop operational safety. Historically, adversarial robustness research has centered on bounded $L_p$-norm noise applied to isolated image classification tasks \cite{mangal2023certifying}. However, such synthetic noise models fail to capture the physical realities of autonomous platforms, which routinely encounter continuous spatial transformations \cite{engstrom2019exploring, sharif2018suitability}, including geometric perturbations induced by dynamic maneuvers or lighting variations. Moreover, evaluating the geometric robustness of such multi-stage perception chains remains an open challenge; existing validation frameworks predominantly focus on single-stage image classification and fail to account for downstream tasks, such as non-linear geometric solvers.

Aerospace represents one of the most stringent domains for autonomous operations, where safety-critical components require validation with strong prior coverage according to aviation safety guidelines (\cite{torens2022guidelines, cappi2024design}). In this setting, Vision-Based Landing (VBL) systems provide essential navigational redundancy when primary signals—such as GPS or the Instrument Landing System (ILS)—are degraded, jammed, or unavailable. A standard VBL pipeline relies on a two-stage perception-to-estimation architecture: an off-the-shelf neural network detects 2D runway keypoint landmarks, which are subsequently fed into a Perspective-n-Point (PnP) geometric solver to estimate the 6-DoF attitude and position of the aircraft \cite{valentin2024probabilistic}. In such hybrid architectures, small spatial perturbations in early 2D keypoint predictions propagate non-linearly through the geometric solver, potentially inducing large divergence in downstream 3D pose estimation and breaching operational safety envelopes.

In this paper, we evaluate the geometric robustness of an end-to-end VBL pipeline, from simple classification tasks to continuous keypoint regression coupled with a PnP solver. Extending foundational insights by Engstrom et al. \cite{engstrom2019exploring} on spatial robustness, we formulate first-order gradient-based attacks directly over the geometric transformation hyperparameters within a differentiable pose estimation solver. Through this formulation, we observe that first-order gradient search severely underestimates vulnerabilities in pose estimation, even underperforming a naive random sampling baseline.

To overcome the limitations of first-order gradient search, we reformulate pose estimation robustness within the framework of Global Lipschitzian Optimization (GLO). Specifically, we adapt the principles behind GeoRobust \cite{wang2023towards}, originally designed to identify worst-case misclassifications; and evaluate whether its underlying theoretical convergence guarantees hold when applied to pose estimation.

We conduct an empirical benchmark applying camera rotation and contrast transformations to a YOLOv8-Pose model trained on the LARDv2 dataset. Our results demonstrate that GeoRobust rapidly prunes over 80\% of the worst-case transformation search space within just ten iterations, establishing an efficient paradigm for isolating vulnerability zones prior to downstream formal verification. To support reproducible research in certifiable autonomous systems, we open-source our complete framework, encompassing the end-to-end YOLOv8-Pose and PnP landing pipeline, parametric transformation search spaces, and GLO validation suite (\url{https://github.com/gregoiretheau0/Relai_Geometric_Robustness.git}).

\section{Spatial Robustness for Vision Based Landing}
\label{sec:problem_formalization}

Vision-Based Landing (VBL) provides essential navigational redundancy when primary signals such as GPS or Instrument Landing Systems (ILS) are degraded or unavailable. Rather than operating as an isolated image classification model, a real-world VBL pipeline functions as a multi-stage algorithmic perception chain designed to estimate the aircraft attitude relative to the runway. 

\subsection{End-to-End Perception Pipeline}
We model the pipeline as a composite mapping $\mathcal{F}: \mathcal{X} \rightarrow \mathbf{y}_{\text{3D}}$, where $\mathcal{X} \subset \mathbb{R}^{H \times W \times C}$ denotes the space of input aerial images, and $\mathbf{y}_{\text{3D}} \in \mathbb{R}^3$ represents the groundtruth 3D aircraft attitude in the runway coordinate frame.

Let $T: \mathcal{X} \times \boldsymbol{\Theta} \rightarrow \mathcal{X}$ denote a continuous, physically plausible transformation operator parametrized by a bounded vector $\boldsymbol{\theta} \in \boldsymbol{\Theta} \subset \mathbb{R}^d$. 

The pipeline consists of three coupled functional stages:

\begin{enumerate}
    \item \textbf{Neural Candidate Detection ($g_{\text{detector}}$):} Given the transformed image $\mathbf{x}' = T(\mathbf{x}; \boldsymbol{\theta})$, the keypoint detector outputs a candidate set of $M$ detected instances:
    \begin{equation}
    g_{\text{detector}}(\mathbf{x}') = \left\{ \big(\mathbf{b}_m, c_m, \mathbf{K}_m\big) \right\}_{m=1}^M
    \end{equation}
    where $\mathbf{b}_m \in \mathbb{R}^4$ represents the bounding box coordinates, $c_m \in [0, 1]$ denotes the detection objectness/confidence score, and $\mathbf{K}_m \in \mathbb{R}^{N \times 2}$ is the regressed 2D coordinate tensor for $N$ runway keypoints. In the litterature, $g_{\text{detector}}$ has been mainly envisaged as single stage object detectors \cite{zouzou2025robust} or transformers \cite{valentin2026mechanistic}.

    \item \textbf{Max-Confidence Keypoint Selection ($\sigma_{\text{select}}$):} To isolate the primary runway target, a selection operator $\sigma_{\text{select}}$ filters candidates and selects the keypoint tensor corresponding to the highest confidence score above an operational threshold $\tau_{\text{conf}}$:
    \begin{equation}
    m^* = \arg\max_{m \in \{1, \dots, M\}} c_m \quad \text{s.t.} \quad c_{m^*} \ge \tau_{\text{conf}}
    \end{equation}
    The selected 2D keypoint matrix is then assigned as $\hat{\mathbf{k}} = \mathbf{K}_{m^*}$. 
    
    \item \textbf{Geometric Perspective-n-Point Solver ($f_{\text{PnP}}$):} The selected 2D keypoints $\hat{\mathbf{k}}$ are paired with known 3D runway geo-coordinates $\mathbf{K}_{\text{3D}} \in \mathbb{R}^{N \times 3}$ and passed to the PnP solver, to compute the 3D position estimate \cite{kim2021vision, parasuraman2026novel}:
    \begin{equation}
    \hat{\mathbf{y}}_{\text{3D}}(\boldsymbol{\theta}) = f_{\text{PnP}}\Big(\sigma_{\text{select}}\big(g_{\text{detector}}(T(\mathbf{x}; \boldsymbol{\theta}))\big), \, \mathbf{K}_{\text{3D}}\Big)
    \end{equation}
\end{enumerate}


\subsection{Bounded Pose Estimation Robustness Problem}

We denote the deviation error as the euclidean translation error $\mathcal{L}_{\text{E2E}}: \boldsymbol{\Theta} \rightarrow \mathbb{R}_{\ge 0}$ between the estimated position and ground-truth 3D position:
\begin{equation}
\mathcal{L}_{\text{E2E}}(\boldsymbol{\theta}) = \left\| \hat{\mathbf{y}}_{\text{3D}}(\boldsymbol{\theta}) - \mathbf{y}_{\text{3D}} \right\|_2
\end{equation}

\begin{definition}[\textbf{Bounded Pose Estimation Robustness}]
Given a nominal image $\mathbf{x} \in \mathcal{X}$, ground-truth position $\mathbf{y}_{\text{3D}} \in \mathbb{R}^3$, bounded parameter domain $\boldsymbol{\Theta}$, and operational safety threshold $\epsilon_{\text{safe}} > 0$, the end-to-end system $\mathcal{F}$ is defined as \emph{$\epsilon_{\text{safe}}$-robust} on $(\mathbf{x}, \boldsymbol{\Theta})$ if and only if:
\begin{equation}
\max_{\boldsymbol{\theta} \in \boldsymbol{\Theta}} \mathcal{L}_{\text{E2E}}(\boldsymbol{\theta}) \le \epsilon_{\text{safe}}
\end{equation}
Conversely, an operational safety breach is triggered if there exists any parameter configuration $\boldsymbol{\theta}^* \in \boldsymbol{\Theta}$ such that:
\begin{equation}
\mathcal{L}_{\text{E2E}}(\boldsymbol{\theta}^*) > \epsilon_{\text{safe}}
\end{equation}
\end{definition}

In our evaluation, we set the critical safety boundary to $\epsilon_{\text{safe}} = 1000\,\text{m}$ and confidence score to $0.25$. Isolating worst-case operational failure modes reduces to solving the global non-convex optimization problem:
\begin{equation}
\boldsymbol{\theta}^* = \arg\max_{\boldsymbol{\theta} \in \boldsymbol{\Theta}} \mathcal{L}_{\text{E2E}}(\boldsymbol{\theta})
\end{equation}

\section{Geometric Robustness for Pose Estimation}
\label{sec:methodology}


\subsection{White-Box Gradient Baseline: APGD Extension}
\label{subsec:first_order_attacks}

To establish a gradient-based white-box baseline, we adapt AutoAttack \cite{croce2020reliable}, the benchmark standard for evaluating $L_p$-norm adversarial robustness. AutoAttack originally comprises an ensemble of Auto Projected Gradient Descent (APGD), Fast Adaptive Boundary (FAB), and Square Attack \cite{croce2020reliable}. In our spatial setting, we exclude Square Attack—since localized patch insertions fail to reflect global camera motion—as well as FAB, whose boundary-projection mechanism is designed for discrete classification decisions rather than continuous 3D pose regression. 

\subsubsection{Enabling End-to-End Differentiability via BPnP}
Standard operational VBL pipelines rely on non-differentiable numerical algorithms (such as OpenCV's iterative PnP) to solve the Perspective-$n$-Point problem. However, evaluating the pipeline's adversarial robustness via first-order methods requires an unbroken gradient flow. To enable gradient-based attacks, we substitute the standard iterative solver during white-box optimization with BPnP \cite{BPnP2020}, an end-to-end backpropagatable PnP module. 

Rather than optimizing intermediate 2D keypoint losses, our adapted APGD directly maximizes the 3D translation error $\mathcal{L}_{\text{E2E}}(\boldsymbol{\theta})$ to uncover operational safety breaches where $\mathcal{L}_{\text{E2E}}(\boldsymbol{\theta}) > \epsilon_{\text{safe}}$.
Incorporating $f_{\text{BPnP}}$, the end-to-end loss gradient with respect to the transformation parameter vector $\boldsymbol{\theta} \in \boldsymbol{\Theta}$ is evaluated via the chain rule:
\begin{equation}
\frac{\partial \mathcal{L}_{\text{E2E}}}{\partial \boldsymbol{\theta}} = \frac{\partial \mathcal{L}_{\text{E2E}}}{\partial \hat{\mathbf{y}}_{\text{3D}}} \cdot \frac{\partial f_{\text{BPnP}}(\hat{\mathbf{k}}, \mathbf{K}_{\text{3D}})}{\partial \hat{\mathbf{k}}} \cdot \frac{\partial \sigma_{\text{select}}}{\partial g_{\text{detector}}} \cdot \frac{\partial g_{\text{detector}}(\mathbf{x}')}{\partial \mathbf{x}'} \cdot \frac{\partial T(\mathbf{x}; \boldsymbol{\theta})}{\partial \boldsymbol{\theta}}
\end{equation}

\subsection{Global Lipschitzian Optimization}
\label{subsec:georobust_glo}

To overcome the inherent breakdown of first-order gradient attacks for geometric transformation, we integrate GeoRobust \cite{wang2023towards}, a derivative-free robustness analyzer based on the DIRECT (DIviding RECTangles) global optimization algorithm \cite{jones1993lipschitzian}.

GeoRobust operates strictly as a black-box evaluator, it is thus compatible with any PnP algorithm. GeoRobust operates as a global minimization algorithm over $\boldsymbol{\Theta}$. We formulate its objective function $l(\boldsymbol{\theta})$ as the margin to operational safety:
\begin{equation}
l(\boldsymbol{\theta}) = \epsilon_{\text{safe}} - \left\| \hat{\mathbf{y}}_{\text{3D}}(\boldsymbol{\theta}) - \mathbf{y}_{\text{3D}} \right\|_2
\end{equation}
Minimizing $l(\boldsymbol{\theta})$ directly maximizes the 3D translation error $\mathcal{L}_{\text{E2E}}(\boldsymbol{\theta})$. A critical safety breach is triggered as soon as $l(\boldsymbol{\theta}) < 0$, indicating that the 3D position error has exceeded the operational threshold $\epsilon_{\text{safe}} = 1000\,\text{m}$.

\subsubsection{Algorithmic Foundation and Search Space Convergence}
Unlike gradient-ascent heuristics that get trapped in local extrema, GeoRobust systematically partitions the bounded parameter domain $\boldsymbol{\Theta} \subset \mathbb{R}^d$ into hyper-rectangles. At each iteration, it identifies and divides \emph{Potentially Optimal (PO)} sub-regions, namely subspaces that cannot be ruled out from containing the global maximum $\boldsymbol{\theta}^*$ under any valid Lipschitz bound.

\paragraph{Convergence Under Domain Bounds.}
Classical Lipschitzian optimization requires knowing the exact Lipschitz constant $L$ of the objective function. DIRECT relaxes this requirement by implicitly sweeping through all feasible $L \in (0, \infty)$ simultaneously. Under Lipschitzian assumption, DIRECT guarantees sound search space coverage: the set of potentially optimal hyper-rectangles retained at iteration $k$ necessarily contains the global optimum $\boldsymbol{\theta}^*$, guaranteeing asymptotic convergence $\lim_{k \to \infty} \max_{\boldsymbol{\theta} \in \mathcal{P}_k} \mathcal{L}_{\text{E2E}}(\boldsymbol{\theta}) = \mathcal{L}_{\text{E2E}}(\boldsymbol{\theta}^*)$ without requiring gradient access.

\subsubsection{Theoretical Guarantees}
Evaluating the objective function $l(\boldsymbol{\theta})$ introduces two non-smooth modules that complicate Lipschitz continuity:
\begin{enumerate}
    \item \textbf{The Candidate Selection Operator ($\sigma_{\text{select}}$):} The discrete $\arg\max$ operation over confidence scores $c_m$ introduces step-wise jump discontinuities when a perturbation causes a different detection box to exceed $\tau_{\text{conf}}$ and become the top candidate. Thus our framework do not cover shifted transformation such as translation.
    \item \textbf{The Geometric PnP Solver ($f_{\text{PnP}}$):} Non-linear 2D-to-3D reprojection can exhibit severe local ill-conditioning near degenerate geometric configurations (e.g., co-planar keypoints or near-zero focal lengths).
\end{enumerate}

\begin{property}[\textbf{Global Lipschitz Continuity of the Hybrid Pipeline}]
\label{prop:global_lipschitz}
Let $\mathcal{X}$ be the space of input images, and let $\boldsymbol{\Theta} \subset \mathbb{R}^d$ be a compact, convex domain of transformation parameters. Consider the end-to-end evaluation map $\mathcal{L}_{\text{E2E}}: \boldsymbol{\Theta} \to \mathbb{R}_{\ge 0}$ defined as the composite mapping:
\begin{equation}
    \mathcal{L}_{\text{E2E}}(\boldsymbol{\theta}) = \left( f_{\text{loss}} \circ f_{\text{PnP}} \circ \hat{\mathbf{k}}_{\sigma_{\text{select}}} \right) (\boldsymbol{\theta}; \mathbf{x})
\end{equation}
Under operational safety conditions ($\epsilon_{\text{safe}}$-robustness), if the following conditions hold:
\begin{enumerate}
    \item \textbf{Invariant Selection:} The candidate selection operator $\sigma_{\text{select}}$ yields a constant primary candidate index $m^*(\boldsymbol{\theta}) = m^*$, removing discrete switching discontinuities on $\boldsymbol{\Theta}$.
    \item \textbf{Lipschitz Keypoint Extraction:} The predicted keypoint mapping $\boldsymbol{\theta} \mapsto \hat{\mathbf{k}}(\boldsymbol{\theta})$ is continuous and globally Lipschitz continuous on $\boldsymbol{\Theta}$ with constant $L_k > 0$ (which holds for Yolo architecture).
    \item \textbf{Non-Degenerate PnP Solver:} The keypoint configurations $\hat{\mathbf{k}}(\boldsymbol{\theta})$ remain strictly non-degenerate, such that the PnP solver $f_{\text{PnP}}$ is continuously differentiable ($\mathcal{C}^1$) via the Implicit Function Theorem \cite{BPnP2020}.
\end{enumerate}
Then, the following properties are satisfied:
\begin{enumerate}
    \item The PnP mapping $f_{\text{PnP}}$ is globally Lipschitz continuous on the compact set of realized non-degenerate keypoint matrices with constant $L_{\text{PnP}} > 0$.
    \item The composite end-to-end error function $\mathcal{L}_{\text{E2E}}(\boldsymbol{\theta})$ is globally Lipschitz continuous on $\boldsymbol{\Theta}$, with Lipschitz constant:
    \begin{equation}
        L_{\text{E2E}} \le L_{\text{loss}} \cdot L_{\text{PnP}} \cdot L_k
    \end{equation}
    thereby guaranteeing the theoretical convergence of Lipschitzian global optimization algorithms (e.g., DIRECT).
\end{enumerate}
\end{property}
\section{Experimental Setup \& Protocols}

\subsection{Dataset and Vision Model Architecture}
To evaluate the geometric robustness of the VBL system, we utilize the Landing Approach Runway Detection (LARD, \cite{wang2024valnet}) dataset, which comprises high-quality aerial images of runways captured during approach and landing phases \cite{ducoffe2023lard}. Specifically, we conduct our benchmarks on the LARDv2 dataset, which provides challenging operational variations in viewpoint, lighting, and weather conditions. 

For the vision component of the VBL pipeline, we employ a state-of-the-art YOLOv8-Pose model trained on LARDv2. YOLOv8-Pose serves as a representative keypoint detection backbone, predicting bounding boxes and regressing the four 2D runway corners required for Perspective-n-Point pose estimation.

\paragraph{Experimental Focus}
While several geometric and photometric transformations satisfy property~\ref{prop:global_lipschitz}, we deliberately restrict our evaluation space to \textbf{camera roll rotation} ($\phi$) and \textbf{photometric contrast scaling} ($\alpha$). This focused subset directly isolates critical in-flight operational hazards: camera roll accurately models aircraft bank maneuvers and atmospheric turbulence, while contrast variation emulates environmental lighting shifts such as cloud cover or sun glare during final approach.

\paragraph{Instantiation of Transformation Parameters}
The general perturbation vector $\boldsymbol{\theta} \in \Theta$ is instantiated according to the evaluation dimensionality:
\begin{itemize}
    \item \textbf{1D Spatial Evaluation:} $\boldsymbol{\theta} = [\phi]^\top$, with continuous roll angle $\phi \in [-\phi_{\text{max}}, \phi_{\text{max}}]$.
    \item \textbf{2D Multi-Dimensional Evaluation:} $\boldsymbol{\theta} = [\phi, \alpha]^\top$, combining roll angle $\phi$ with contrast factor $\alpha \in [\alpha_{\text{min}}, \alpha_{\text{max}}]$.
\end{itemize}

To prevent artificial boundary artifacts (such as zero-padded black borders), we employ reflection padding.

\subsection{Evaluation Protocols and Operational Budgets}
To comprehensively assess the pipeline's resilience against the aforementioned 1D and 2D perturbations, we designed two distinct evaluation protocols:

\begin{itemize}
    \item \textbf{Spatial Budget Evaluation:} We allocate a restricted budget of permissible rotation angles, evaluating the specific interval $[2^{\circ}, 5^{\circ}, 10^{\circ}, 15^{\circ}, 20^{\circ}, 30^{\circ}]$. This range is strictly grounded in flight dynamics: $0^{\circ}$ to $5^{\circ}$ represents standard crosswinds and light turbulence, which the system must resist to. The $10^{\circ}$ to $15^{\circ}$ range simulates heavy turbulence and tight approach maneuvers, a key threshold where traditional artificial horizon-based models may begin to drift. The upper bounds of $20^{\circ}$ to $30^{\circ}$ represent extreme conditions (\eg, severe wind shear) or evasive maneuvers. Because commercial airliners rarely exceed a $30^{\circ}$ bank angle during standard operations, this interval allows us to map robustness up to the physical limit. 
    
    \item \textbf{Runtime Budget Evaluation:} In contrast to spatial bounds, we introduce a strict computational constraint by allocating a fixed time budget ($n \in [0.1, 0.5, 1.0, 2.0, 5.0]$ seconds) per image. 
    
\end{itemize}

\subsection{Evaluation Metrics and Optimization Tracking}
\label{subsec:metrics}

We consider the following metrics:

\begin{itemize}
    \item \textbf{Median Deviation Error ($\tilde{\mathcal{L}}_{\text{E2E}}$):} The median of the deviation error $\mathcal{L}_{\text{E2E}}(\boldsymbol{\theta})$ across the evaluation set.

    \item \textbf{Survival Rate ($S_{\text{rate}}$):} The percentage of test samples maintaining pose estimation within the operational safety threshold $\epsilon_{\text{safe}}= 1000 \, \mathrm{m}$:
    \begin{equation}
        S_{\text{rate}}(\boldsymbol{\theta}) = \frac{1}{N} \sum_{i=1}^{N} \mathbb{I}\left( \mathcal{L}_{\text{E2E}}^{(i)}(\boldsymbol{\theta}) \le \epsilon_{\text{safe}} \right) \times 100\%,
    \end{equation}
    where $\mathbb{I}(\cdot)$ denotes the indicator function and $N$ is the total sample count.

    \item \textbf{Remaining Search Area ($\mathrm{RA}$):} The normalized hyper-volume of candidate transformation domains enclosing potential worst-case perturbations at iteration $k$:
    \begin{equation}
        \mathrm{RA}(k) = \frac{\sum_{i \in \mathrm{PO}_k} \mathrm{Vol}(S_i)}{\mathrm{Vol}(\mathcal{S})} \times 100\%,
    \end{equation}
    where $\mathrm{PO}_k$ is the set of active \textit{Potentially Optimal} (PO) hyper-rectangles retained by DIRECT partitioning at iteration $k$, $\mathrm{Vol}(S_i)$ is their individual volume, and $\mathrm{Vol}(\mathcal{S})$ is the total volume of the initial search domain.

    \item \textbf{Active PO Subspaces ($|\mathrm{PO}_k|$):} The cardinality of active potentially optimal hyper-rectangles evaluated at step $k$, characterizing algorithm branching efficiency and GPU parallelization load.

    \item \textbf{Critical Perturbation Threshold ($\boldsymbol{\theta}_{\text{crit}}$ / $\phi_{\text{crit}}$):} The minimum perturbation magnitude required to drop the pipeline survival rate below $10\%$ ($S_{\text{rate}} < 10\%$). For 1D parametric evaluations, $\phi_{\text{crit}}$ specifically denotes the critical roll threshold.
\end{itemize}

\section{Experimental Results and Robustness Analysis}

In this section, we present a comprehensive evaluation of the Vision-Based Landing (VBL) system against plausible perturbations. To accurately reflect in-flight hazards, we shift away from standard pixel-wise noise ($L_p$) toward geometric and environmental transformations. The evaluations are conducted on a subset of 100 randomly sampled images from the LARDv2 dataset, preceded by a solver validation across 1,000 images.

\subsection{Validation of the Differentiable Solver (BPnP)}
To ensure that adversarial evaluation metrics are not biased by the choice of the geometric solver, we first evaluate the equivalence between OpenCV's standard non-differentiable Iterative PnP ($f_{\text{IterPnP}}$) and the differentiable BPnP module ($f_{\text{BPnP}}$) used for gradient-based attacks. We conduct a comparative analysis across 1,000 images under two regimes: using ground-truth 2D keypoints and noisy YOLOv8-Pose predictions.

\begin{figure}[tb]
    \centering
    \includegraphics[width=0.8\linewidth]{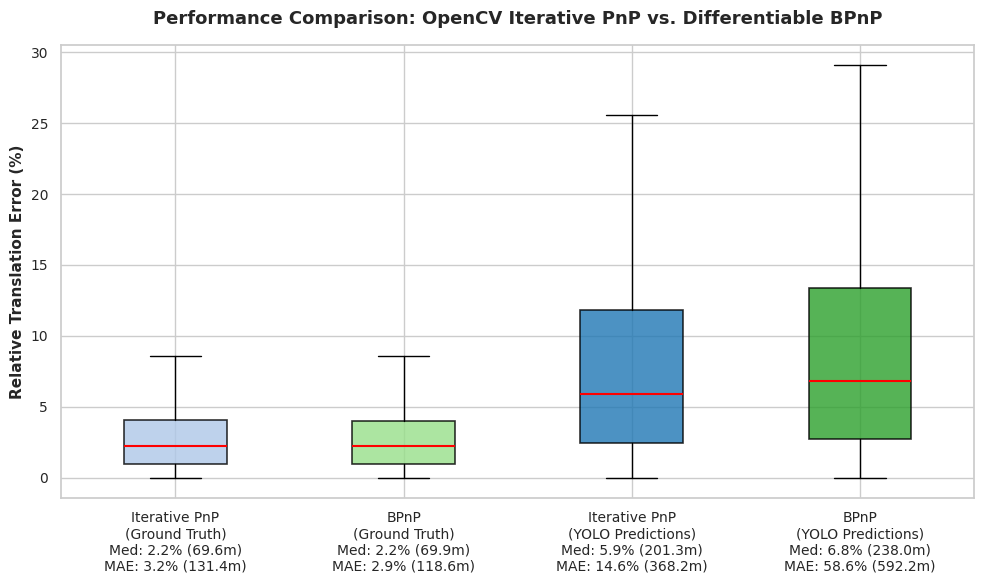}
    \caption{\textbf{Performance Comparison between Iterative PnP and BPnP.} Relative 3D translation errors evaluated on 1,000 LARDv2 images using ground-truth keypoints and YOLOv8-Pose predictions.}
    \label{fig:pnp_solver_validation}
\end{figure}

As shown in \cref{fig:pnp_solver_validation}, under nominal noise-free conditions (Ground Truth keypoints), BPnP demonstrates near-perfect fidelity to standard Iterative PnP, achieving an identical median relative translation error of $2.2\%$ ($\approx 69.6\text{m}$ for Iterative PnP vs. $69.9\text{m}$ for BPnP) and comparable Mean Absolute Error (MAE of $3.2\%$ vs. $2.9\%$). When evaluated on noisy keypoint predictions from YOLOv8-Pose, both solvers maintain close median error profiles ($5.9\%$ / $201.3\text{m}$ for Iterative PnP vs. $6.8\%$ / $238.0\text{m}$ for BPnP). Although BPnP exhibits higher variance under severe prediction outliers (MAE of $58.6\%$), its baseline consistency confirms that BPnP acts as a faithful differentiable surrogate. This validates that the performance degradation of gradient-based attacks (APGD) observed in subsequent experiments is caused by non-convex loss landscapes rather than inaccuracies inherent to the differentiable PnP solver.

\subsection{Spatial Robustness to 1D Camera Rotations}
Our initial experiment evaluates the pipeline's resilience against 1D spatial perturbations, specifically camera roll rotations simulating aircraft turbulence. We expand the budget of permissible rotation angles, spanning from standard turbulence ($\pm 2^{\circ}$) up to the physical limits of evasive maneuvers ($\pm 30^{\circ}$), and evaluate the pipeline Survival Rate ($S_{\text{rate}}$).

\begin{figure}[t]
  \centering
  \begin{subfigure}[b]{0.49\textwidth}
    \centering
    \includegraphics[width=\textwidth]{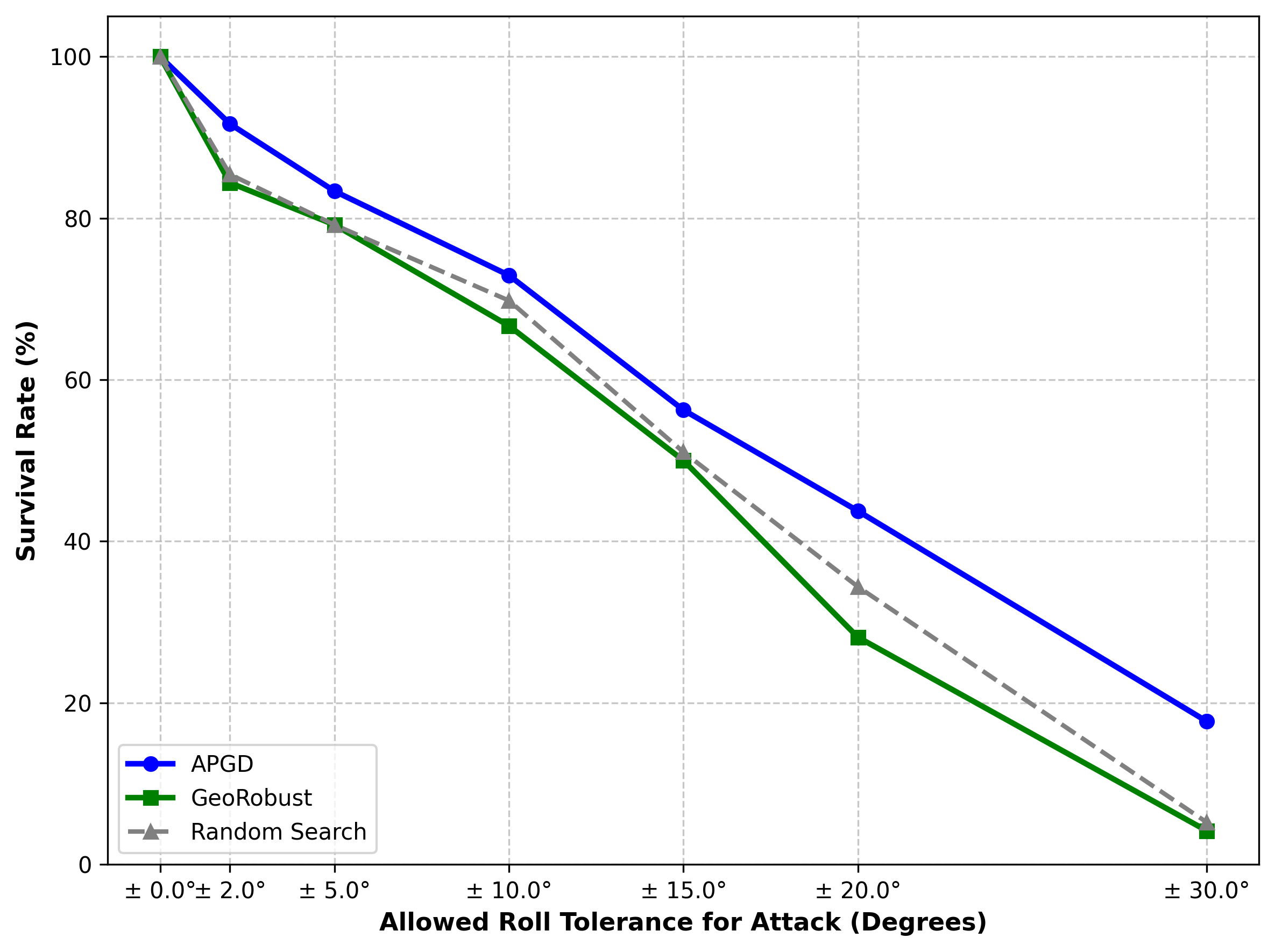} 
    \caption{1D Camera Rotations}
    \label{fig:spatial_1d}
  \end{subfigure}
  \hfill
  \begin{subfigure}[b]{0.49\textwidth}
    \centering
    \includegraphics[width=\textwidth]{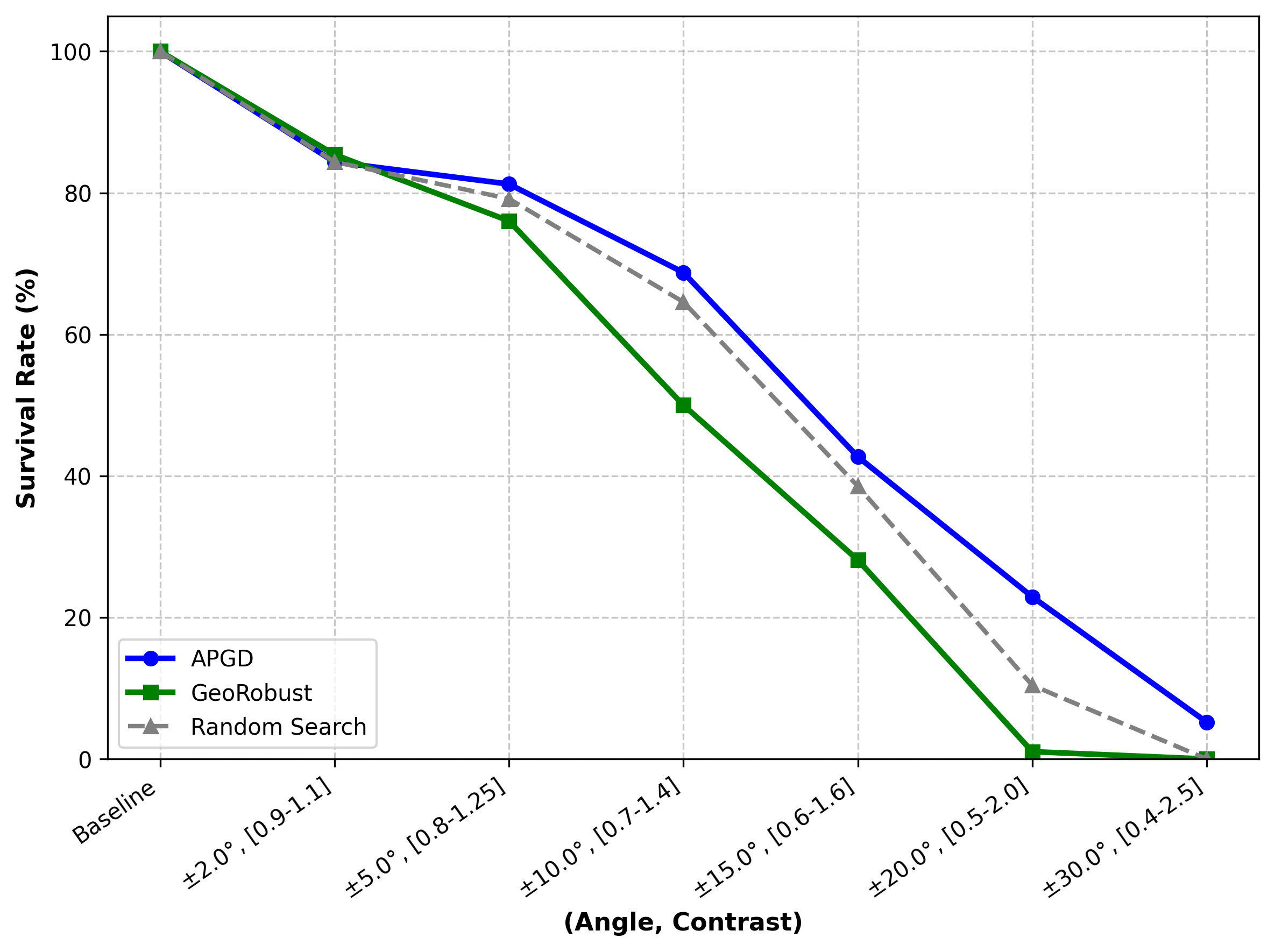} 
    \caption{2D Rotation \& Contrast}
    \label{fig:spatial_2d}
  \end{subfigure}
  \caption{Spatial Robustness Analysis. \textbf{(a)} Survival rate across increasing 1D roll rotation budgets ($\pm2^\circ$ to $\pm30^\circ$). \textbf{(b)} Survival rate under compounding 2D rotation and contrast tolerances.}
  \label{fig:spatial_robustness}
\end{figure}

As shown in \cref{fig:spatial_1d}, we compare three distinct optimization approaches: AutoAttack (APGD, a white-box gradient method), GeoRobust (a black-box global optimizer), and a Random Search baseline. The gradient-based APGD attack struggles to optimize over the highly non-convex 1D transformation space, causing APGD to stagnate in local minima and fail to outperform the Random Search baseline. Conversely, GeoRobust leverages global Lipschitzian partitioning to navigate this non-smooth landscape, consistently identifying worst-case geometric perturbations and achieving the lowest survival rate across all angular budgets. This highlights that first-order gradient information is often insufficient for geometric robustness auditing, reinforcing the necessity of global sampling strategies like GeoRobust. 

\subsection{Runtime Robustness and Attack Efficiency}
In an operational setting, the computational time required to discover an adversarial perturbation is a key evaluation metric for understanding real-world threat plausibility. In this second experiment, we constrain the spatial rotation budget to a fixed, wide interval ($\pm 15^{\circ}$) and allocate varying computational time budgets (ranging from $0.1$s to $5.0$s per image) to mirror scenarios ranging from instantaneous intra-frame vulnerabilities to theoretical worst-case optimization convergence.

\begin{figure}[tb]
    \centering
    \begin{subfigure}[b]{0.48\linewidth}
        \centering
        \includegraphics[width=\linewidth]{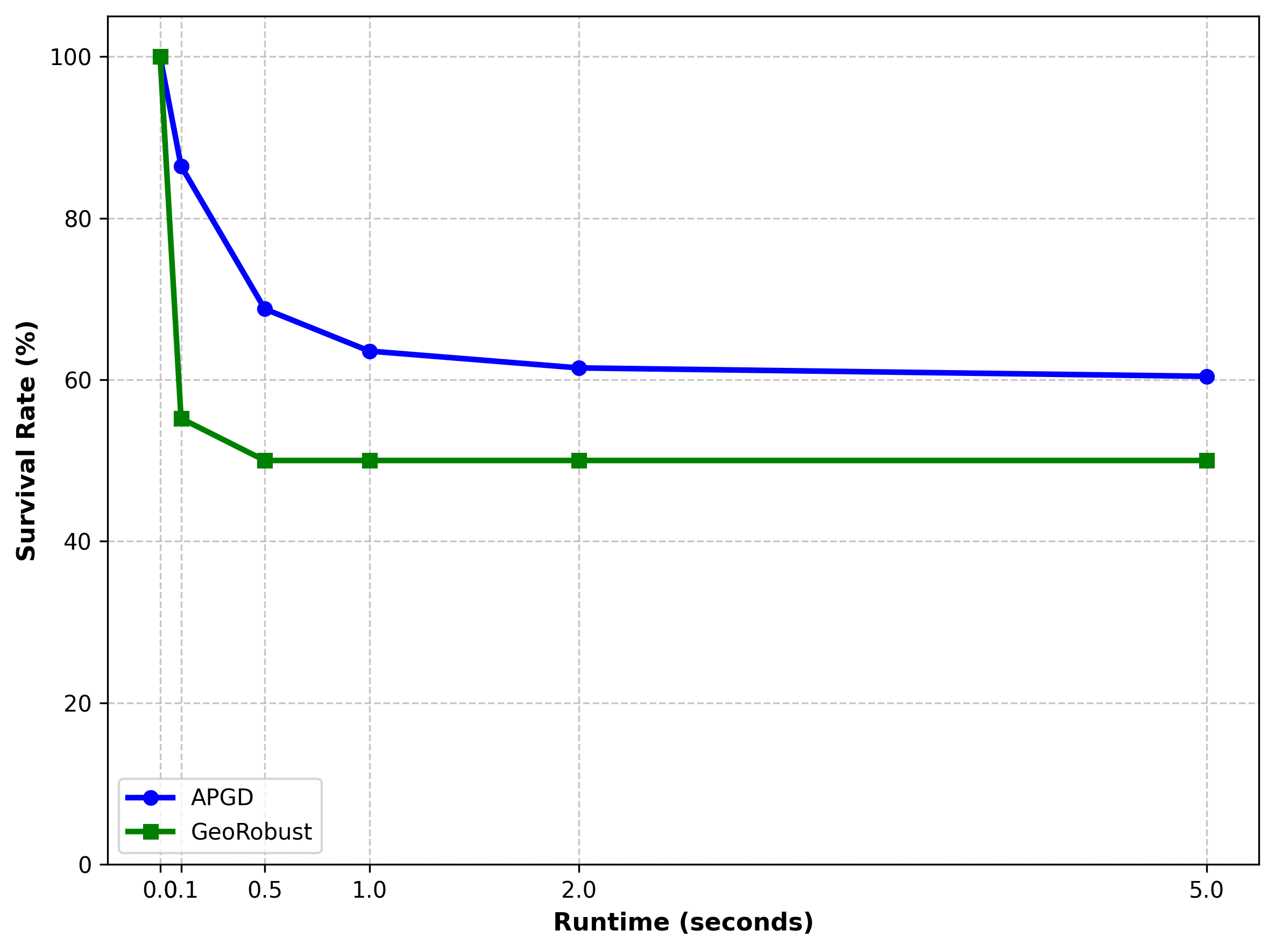}
    \end{subfigure}
    \hfill
    \begin{subfigure}[b]{0.48\linewidth}
        \centering
        \includegraphics[width=\linewidth]{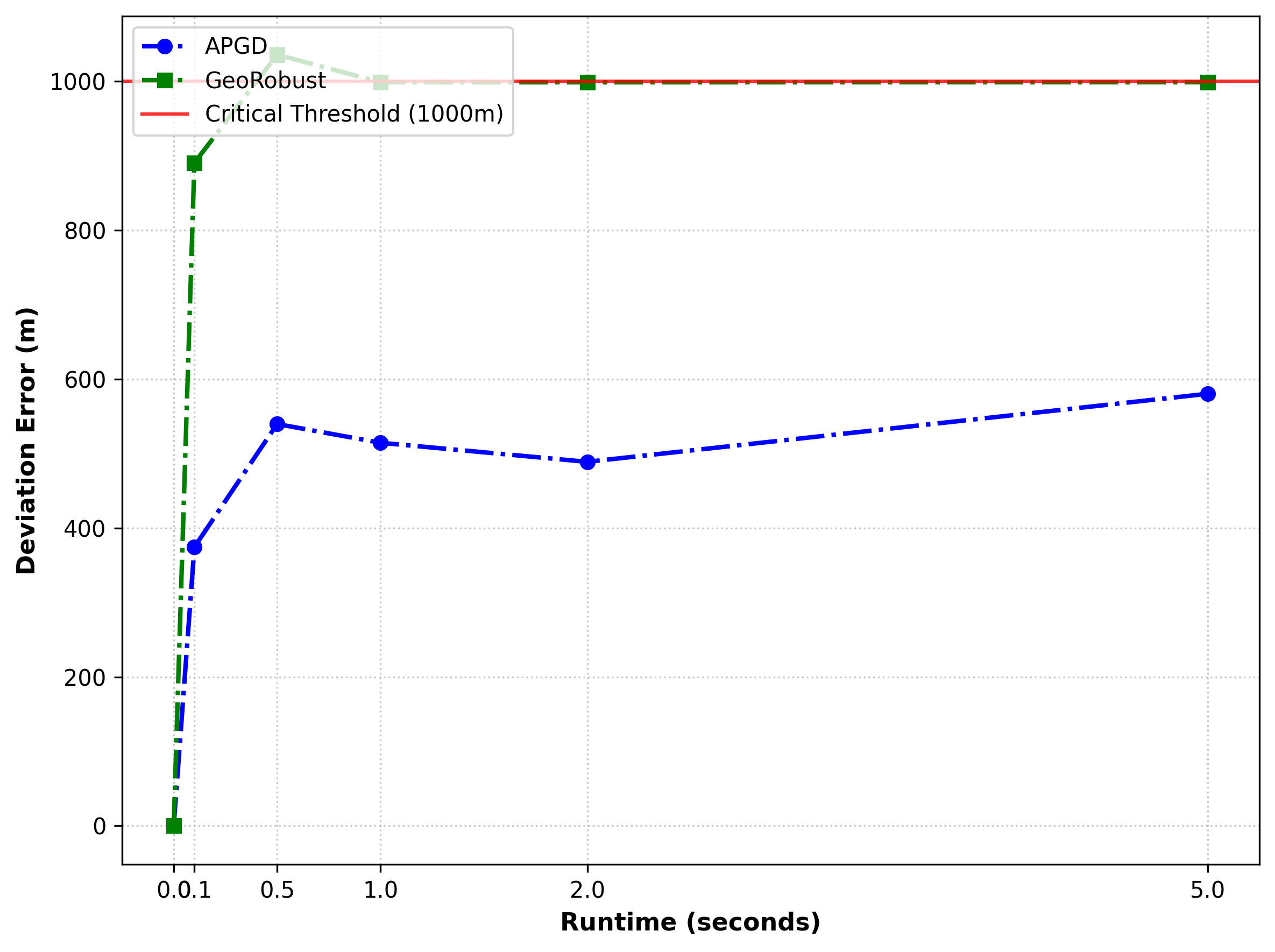}
    \end{subfigure}
    \caption{\textbf{Runtime Attack Efficiency.} Impact of strict computational time budgets ($0.1$s to $5.0$s) on survival rate and median translation error under a fixed $\pm 15^{\circ}$ rotation bound.}
    \label{fig:temporal_robustness}
\end{figure}

By tracking both $S_{\text{rate}}$ and the median deviation error ($\tilde{\mathcal{L}}_{\text{E2E}}$) over time, we evaluate the real-time efficiency of the attacks. As illustrated in \cref{fig:temporal_robustness}, the runtime statistics reveal the convergence behaviors of APGD and GeoRobust: while both algorithms successfully exploit the spatial bounds to degrade system accuracy, their time-to-failure rates differ, illustrating how quickly spatial divergence can be synthesized under strict time constraints.

\subsection{Multi-Dimensional Perturbations: Rotation and Contrast}
To more closely approximate compounding real-world environmental conditions, we extend our robustness evaluation to a 2D perturbation space, combining geometric transformations (rotation) with photometric alterations (contrast). 

\begin{figure}[tb]
    \centering
    \begin{subfigure}[b]{0.48\linewidth}
        \centering
        \includegraphics[width=\linewidth]{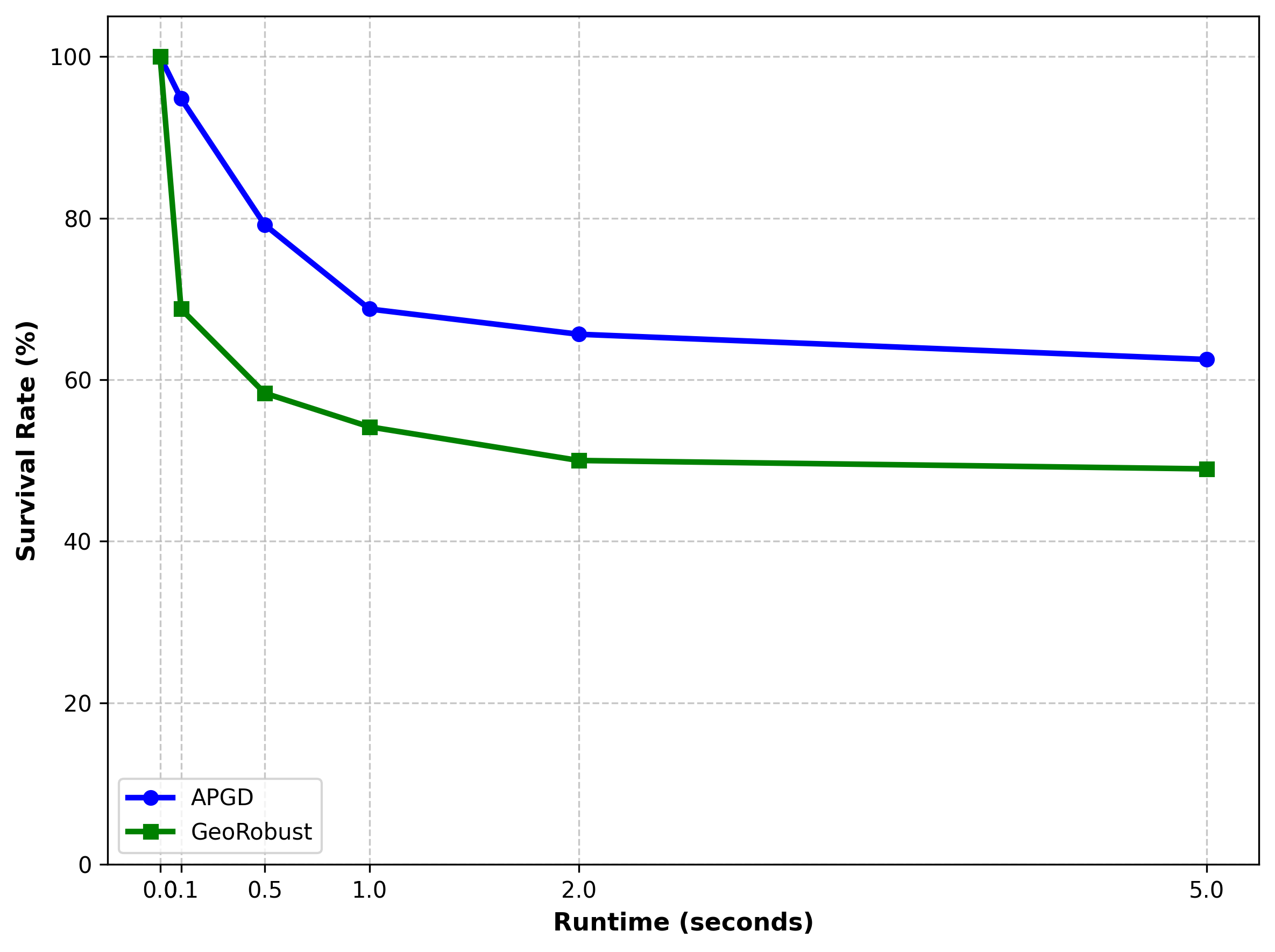}
    \end{subfigure}
    \hfill
    \begin{subfigure}[b]{0.48\linewidth}
        \centering
        \includegraphics[width=\linewidth]{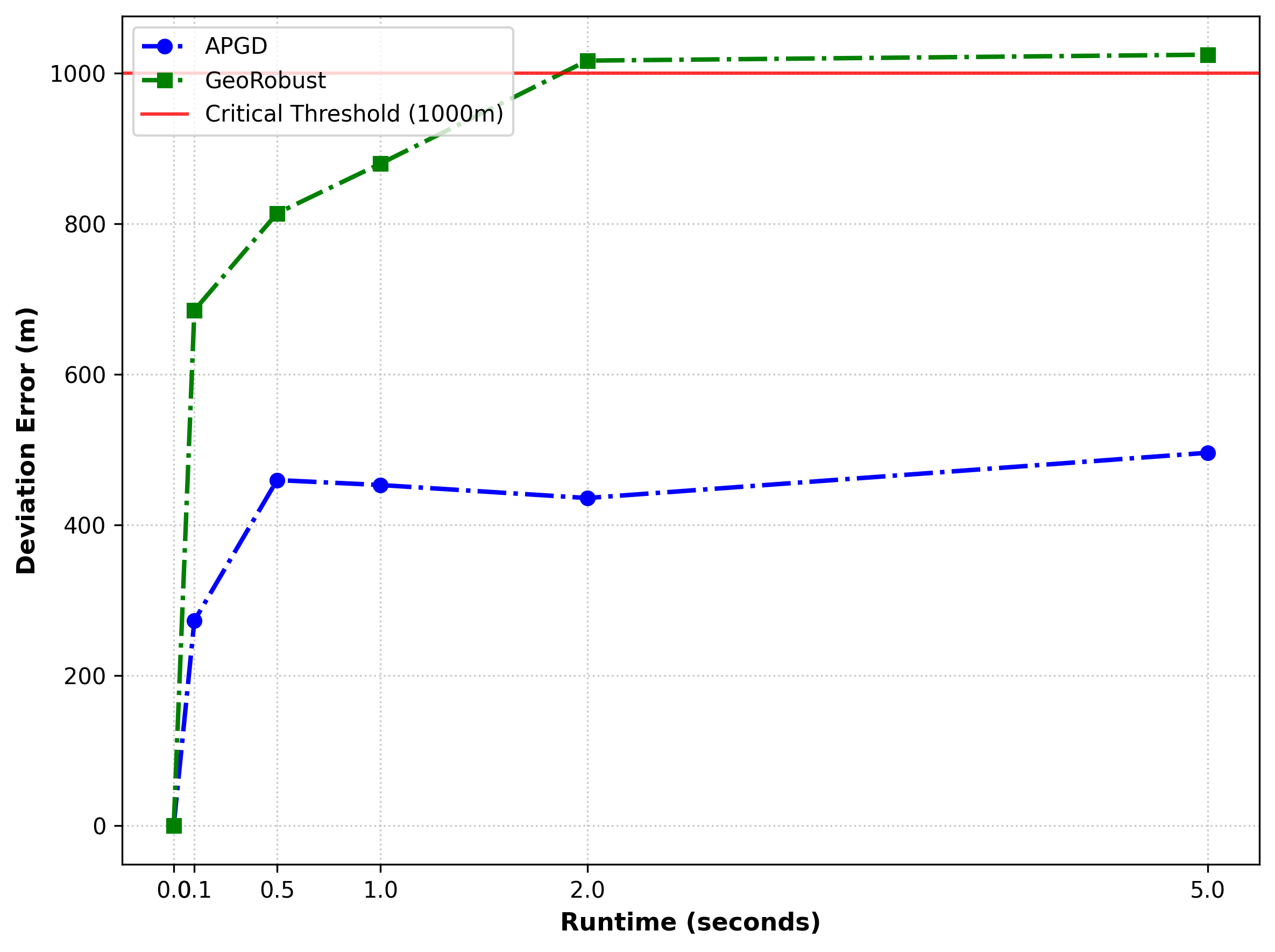}
    \end{subfigure}
    \caption{\textbf{Runtime Attack Efficiency under 2D Perturbations.} Impact of computational time budgets (0.1s to 5.0s) on Survival rate and median translation error under fixed $\pm 10.0^{\circ}$ rotation and $[0.7, 1.4]$ contrast bounds.}
    \label{fig:temporal_robustness_2d}
\end{figure}

\begin{itemize}
    \item \textbf{Spatial Budget Evaluation (2D):} Following the methodology of the 1D evaluation, we incrementally expand the combined tolerances. The constraints scale from standard turbulence with minor lighting changes (\eg, $\pm 2.0^{\circ}$, contrast $[0.9, 1.1]$) up to extreme conditions (\eg, $\pm 30.0^{\circ}$, contrast $[0.4, 2.5]$). As illustrated in \cref{fig:spatial_2d}, introducing a second dimension increases the non-convexity of the optimization space. Consequently, the gradient-based APGD struggles significantly, falling behind the Random Search baseline. GeoRobust, however, maintains its efficiency in isolating multi-dimensional adversarial examples, successfully driving the survival rate to nearly 0\% under extreme conditions.
    
    \item \textbf{Runtime Budget Evaluation (2D):} We subsequently fix the 2D perturbation limits to a realistic operational hazard zone (rotation $\pm 10.0^{\circ}$, contrast $[0.7, 1.4]$) and apply strict time budgets. As shown in \cref{fig:temporal_robustness_2d}, the runtime statistics demonstrate a clear divergence in optimization capabilities: while APGD's median deviation error plateaus around $500\text{m}$ due to entrapment in local extrema, GeoRobust efficiently navigates the non-convex 2D landscape, crossing the $1000\text{m}$ failure threshold within just $2.0\text{s}$.
\end{itemize}

\subsection{Vulnerability Profiling by Approach Distance}
During a flight approach, the impact of a geometric perturbation heavily depends on the aircraft's distance to the runway. To quantify this, we categorize the approach images into quartiles based on their ground-truth slant distance. For each distance interval, we measure the Critical Roll Tolerance ($\phi_{\text{crit}}$) required to drop the pipeline survival rate below $10\%$.

\begin{figure}[tb]
    \centering
    \includegraphics[width=0.82\linewidth]{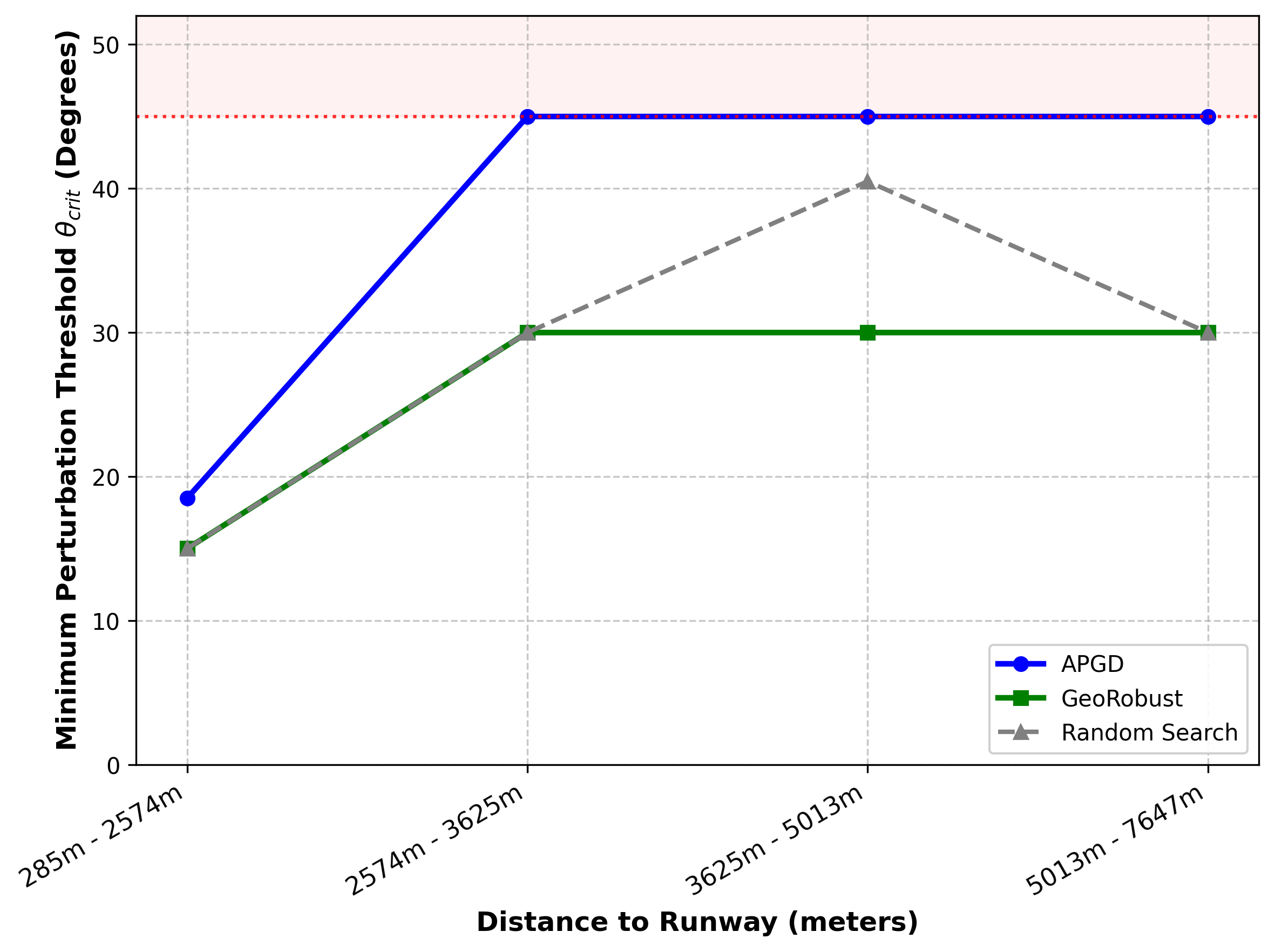}
    \caption{\textbf{Distance-Based Vulnerability Profile.} The roll angle required to achieve a 90\% failure rate across different approach distances (divided into quartiles). The red shaded region ($\geq 45^{\circ}$) denotes the experimental limit: attacks reaching this boundary fail to compromise the pipeline within physically plausible rotations, establishing a guaranteed robustness zone.}
    \label{fig:vulnerability_profile}
\end{figure}

As depicted in \cref{fig:vulnerability_profile}, the system's robustness varies significantly across the approach phases. At close range (\eg, under 2600m), the VBL system is sensitive to geometric shifts, requiring only minor roll perturbations (around $15^{\circ}$) for GeoRobust to systematically compromise the downstream pose estimation. Conversely, as the distance increases, the required angular perturbation scales up, stabilizing around $30^{\circ}$ for mid-to-long range approaches. 
\paragraph{Handling Total Detection Discontinuities}
If an adversarial perturbation causes the 2D vision model to completely miss the runway or output fewer than the four keypoints required by the PnP algorithm, the downstream 3D pose solver cannot execute. To handle this discontinuity, any total detection breakdown is assigned a default penalty translation error of $\mathcal{L}_{\text{E2E}} = 5000\text{m}$.

Consistent with our previous findings, the gradient-based APGD method underperforms in this geometric context. Across most distance intervals, APGD fails to identify vulnerabilities, plateauing at the $45^{\circ}$ experimental limit, our defined boundary for guaranteed robustness, without breaching the error threshold. 

This experiment yields a comprehensive distance-based vulnerability profile, mapping out exactly how much physical disturbance the VBL system can withstand at different phases of the landing approach. Quantifying these exact failure envelopes demonstrates the necessity of black-box global optimizers like GeoRobust and provides a benchmark for future aeronautical certification processes.

\subsection{Convergence Analysis and Search Space Reduction}
To gain deeper insights into the internal mechanics of the global Lipschitzian optimization, we track the evolution of GeoRobust's search space over 30 iterations under a fixed $\pm 15^{\circ}$ rotation budget across the 100 test images. Unlike local gradient-based methods, GeoRobust systematically partitions the transformation space into candidate hyper-rectangles.

\begin{figure}[tb]
    \centering
    \includegraphics[width=0.8\linewidth]{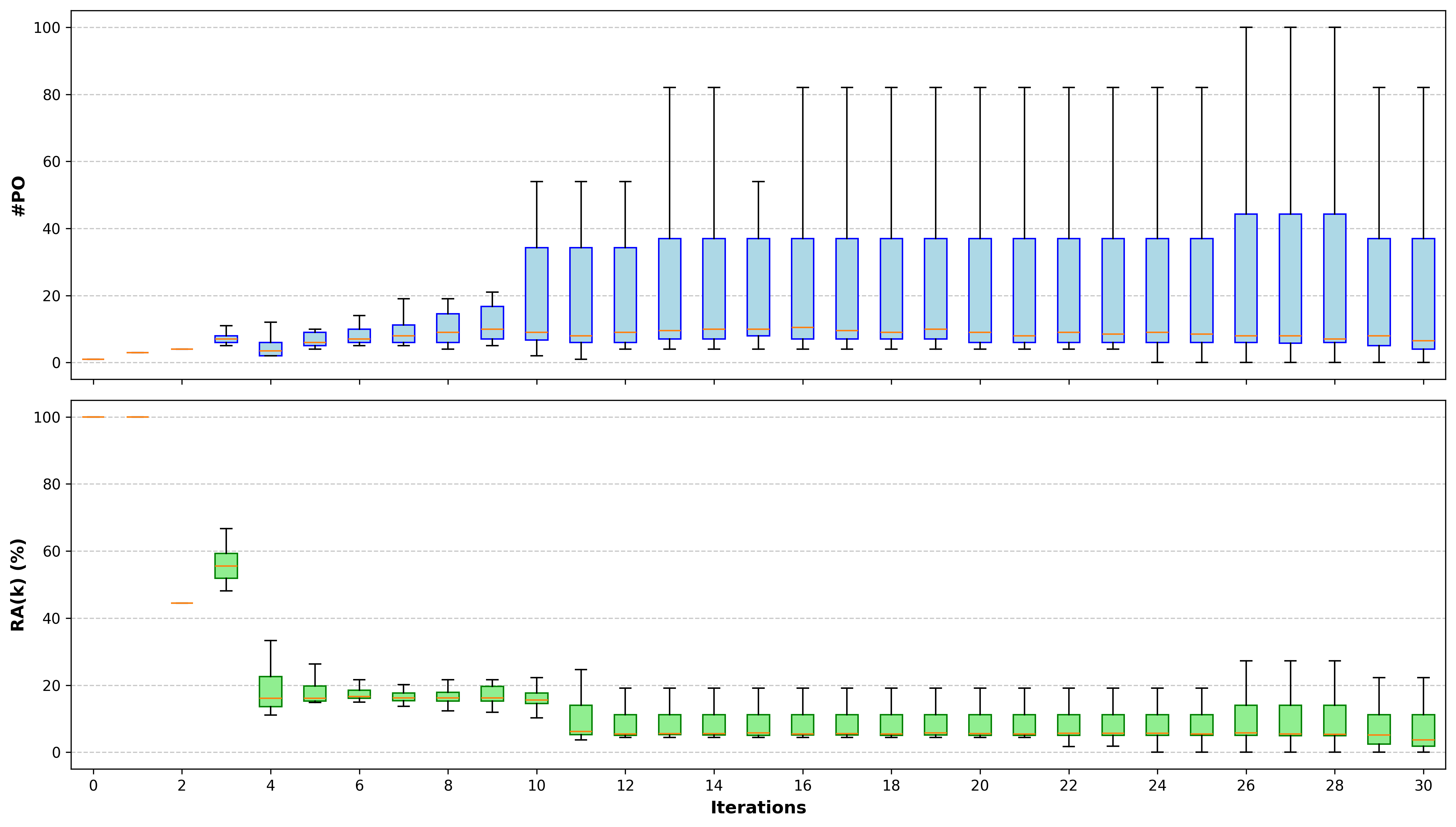}
    \caption{\textbf{GeoRobust Convergence and Search Space Reduction.} Top: count of active PO subspaces ($|\text{PO}_k|$). Bottom: remaining search area ($\text{RA}(k)$) enclosing worst-case transformations.}
    \label{fig:po_convergence}
\end{figure}

As illustrated in \cref{fig:po_convergence}, we monitor the dynamic interplay between Active PO Subspaces ($|\text{PO}_k|$) and the Remaining Search Area ($\text{RA}(k)$). The top plot displays the count of concurrently evaluated candidate spaces. To maximize GPU hardware throughput, GeoRobust adapts standard DIRECT optimization by querying an expanded set of hyper-rectangles simultaneously, which accounts for the initial increase and subsequent stable plateau in $|\text{PO}_k|$.

The bottom plot tracks the spatial collapse of the search area ($\text{RA}$). Starting from full uncertainty ($100\%$ domain coverage at iteration 0), the global Lipschitzian search rapidly discards mathematically safe regions. Within the first 10 iterations, the bounding area enclosing worst-case transformations collapses by over $80\%$, narrowing the critical vulnerability zone to less than $20\%$ of the initial operational bounds.

\section{Conclusion and Future Work}

In this work, we addressed the challenge of Robust Validation in safety-critical aerospace applications. By modeling an end-to-end Vision-Based Landing (VBL) system combining a YOLOv8-Pose surrogate detector and a Perspective-n-Point (PnP) solver, we established a benchmark for evaluating system resilience against physically plausible geometric and photometric perturbations.

Our comparative analysis revealed fundamental insights into assessing the robustness of autonomous vision systems. Standard gradient-based methods like APGD highlight the inherent limits of first-order attacks, failing to reliably identify spatial vulnerabilities because the highly non-convex nature of geometric transformation landscapes induces local extrema that trap first-order optimization \cite{engstrom2019exploring}, thereby yielding overly optimistic robustness estimates. Conversely, black-box global Lipschitzian optimization (GeoRobust) demonstrates strong efficiency by successfully navigating this complex loss landscape, exposing operational failures ($>1000\text{m}$ translation error) within real-time time budgets ($2.0\text{s}$). Crucially, this global approach opens concrete pathways for certification: GeoRobust rapidly collapses the active search space enclosing worst-case perturbations by over $80\%$ within just 10 iterations. This dynamic reduction provides an essential bridge toward formal aeronautical certification by isolating narrow vulnerability zones that can subsequently be targeted by computationally intensive exact verifiers.

Moving forward, we plan to extend this framework along three key axes: (i) evaluating temporal persistence and video-stream consistency across sequential frames during the final approach, (ii) integrating formal verifiers directly into the bounded search spaces isolated by GeoRobust for hybrid certification, and (iii) deploying our open-source auditing protocol on flight-certified target architectures.
\bibliographystyle{splncs04}
\bibliography{biblio}

\begin{thebibliography}{10}
\providecommand{\url}[1]{\texttt{#1}}
\providecommand{\urlprefix}{URL }
\providecommand{\doi}[1]{https://doi.org/#1}

\bibitem{cappi2024design}
Cappi, C., Cohen, N., Ducoffe, M., Gabreau, C., Gardes, L., Gauffriau, A., Ginestet, J.B., Mamalet, F., Mussot, V., Pagetti, C., et~al.: How to design a dataset compliant with an ml-based system odd? arXiv preprint arXiv:2406.14027  (2024)

\bibitem{BPnP2020}
Chen, B., Parra, A., Cao, J., Li, N., Chin, T.J.: End-to-end learnable geometric vision by backpropagating {PnP} optimization. In: {IEEE/CVF} Conference on Computer Vision and Pattern Recognition ({CVPR}) (2020)

\bibitem{croce2020reliable}
Croce, F., Hein, M.: Reliable evaluation of adversarial robustness with an ensemble of diverse parameter-free attacks. In: International Conference on Machine Learning ({ICML}). pp. 2206--2216. PMLR (2020)

\bibitem{ducoffe2023lard}
Ducoffe, M., Carrere, M., Féliers, L., Gauffriau, A., Mussot, V., Pagetti, C., Sammour, T.: {LARD} -- landing approach runway detection -- dataset for vision based landing (2023)

\bibitem{engstrom2019exploring}
Engstrom, L., Tran, B., Tsipras, D., Schmidt, L., Madry, A.: Exploring the landscape of spatial robustness. In: International Conference on Machine Learning ({ICML}). pp. 1802--1811. PMLR (2019)

\bibitem{jones1993lipschitzian}
Jones, D.R., Perttunen, C.D., Stuckman, B.E.: Lipschitzian optimization without the lipschitz constant. Journal of Optimization Theory and Applications  \textbf{79}(1),  157--181 (1993)

\bibitem{kim2021vision}
Kim, S., Kim, J., Park, J., Lee, D.: Vision-based pose estimation of fixed-wing aircraft using you only look once and perspective-n-points. Journal of Aerospace Information Systems  \textbf{18}(9),  659--664 (2021)

\bibitem{mangal2023certifying}
Mangal, R., Leino, K., Wang, Z., Hu, K., Yu, W., Pasareanu, C., Datta, A., Fredrikson, M.: Is certifying {$\ell_p$} robustness still worthwhile? arXiv preprint arXiv:2310.09361  (2023)

\bibitem{parasuraman2026novel}
Parasuraman, I., Ghosh, S., Kumar, A., Chopra, S.: A novel vision-based pose estimation framework for autonomous landing of fixed-wing aircraft. In: AIAA AVIATION 2026 Forum. p.~4421 (2026)

\bibitem{sharif2018suitability}
Sharif, M., Bauer, L., Reiter, M.K.: On the suitability of lp-norms for creating and preventing adversarial examples. In: Proceedings of the IEEE conference on computer vision and pattern recognition workshops. pp. 1605--1613 (2018)

\bibitem{szegedy2014intriguing}
Szegedy, C., Zaremba, W., Sutskever, I., Bruna, J., Erhan, D., Goodfellow, I., Fergus, R.: Intriguing properties of neural networks. In: The Second International Conference on Learning Representations ({ICLR}) (2014)

\bibitem{torens2022guidelines}
Torens, C., Durak, U., Dauer, J.C.: Guidelines and regulatory framework for machine learning in aviation. In: AIAA scitech 2022 forum. p.~1132 (2022)

\bibitem{valentin2026mechanistic}
Valentin, R., Bruvik, O.B., Schlichting, M.R., Kochenderfer, M.J.: Mechanistic interpretability for learning assurance of a vision-based landing system. arXiv preprint arXiv:2605.20607  (2026)

\bibitem{valentin2024probabilistic}
Valentin, R., Katz, S.M., Lee, J., Walker, D., Sorgenfrei, M., Kochenderfer, M.J.: Probabilistic parameter estimators and calibration metrics for pose estimation from image features. In: 2024 AIAA DATC/IEEE 43rd Digital Avionics Systems Conference (DASC). pp. 01--09. IEEE (2024)

\bibitem{wang2023towards}
Wang, F., Xu, P., Ruan, W., Huang, X.: Towards verifying the geometric robustness of large-scale neural networks. In: Proceedings of the {AAAI} Conference on Artificial Intelligence. vol.~37, pp. 15183--15191 (2023)

\bibitem{wang2024valnet}
Wang, Q., Feng, W., Zhao, H., Liu, B., Lyu, S.: Valnet: Vision-based autonomous landing with airport runway instance segmentation. Remote Sensing  \textbf{16}(12), ~2161 (2024)

\bibitem{zouzou2025robust}
Zouzou, A., And{\'e}ol, L., Ducoffe, M., Boumazouza, R.: Robust vision-based runway detection through conformal prediction and conformal map. In: Conformal and Probabilistic Prediction with Applications. vol.~266, pp. 515--534 (2025)

\end{thebibliography}
\end{document}